\documentclass[conference]{IEEEtran}

\usepackage{cuted}
\usepackage{graphicx}
\usepackage{cite}
\usepackage{amsmath,amssymb,amsfonts}
\usepackage{graphicx}
\usepackage{textcomp}
\usepackage{xcolor}
\usepackage{booktabs}
\usepackage{multirow}
\usepackage{siunitx}
\usepackage{hyperref}
\usepackage{stfloats}
\newcommand{\ea}{\textsuperscript{*}}
\newcommand{\nw}{\textsuperscript{\dag}}
\begin{document}

\title{Tactus: Tactile \& Pose Sensing Glove for Natural Dexterity Data Collection}
%\ \
%\large Design, Calibration and Performance Benchmarking}
\author{%
  \IEEEauthorblockN{%
    \parbox{0.95\textwidth}{%
      \centering{Li Guan\ea,  Rachael Granberry\ea, Sahar Aseeri\ea, Sahar Rostami\ea, Taylor Niehues\ea, Sai Dhanoosh Marri\ea, Mia Huang\ea, Dingtian Zhang\ea, Ziyi Kou\ea, Jason Pursel\ea, Eric Whitmire\ea, Vatsal Mehta\ea,  Ashish Deshpande\ea \nw,  Tristan Trutna\ea, Reed Belcher\nw, Emelia Black, Julian Blake\nw, Roger Boldu, Sean Bonaventure, Einar Boyovich, Jordan Brennan\nw, Elizabeth Casillas\nw, David Christensen, Tyler Christensen, Nicholas Colonnese, Stuart Dealey, Gaige DeHaven, Katie Doran, Chris Doughty, Robert Drury\nw, Silvia Espino, Gabriel Geyne, Beth Goodrich\nw, Kyra Gould, Joe Grisso, Susana Guerrero\nw, Andrew Gwozdziowski, Monica Ha, Ehsan Hajiesmaili, Kate Healy, Rose Hope, Michael Hurst, Claire Ikemiya, Daisy Isibor, Rick James, Yashasvi Joginapally, Samuel Johnson\nw, Tyler Jorgensen\nw, Kristy Jost, Ben Kamath\nw, Slava Karulin\nw, Alexander Keener, Woo Kim\nw, Alison Krauskopf, Ankit Kumar, Joshua Lang, Lizzy Li, Cameron Lynch, Daniel Ma, Hector Martinon, Christina Matos, Keagan McCurdy, Lee Mincey, Stuart Nezlek\nw, Tiffany Nguyen, James O'Hara\nw, Jason Osborne, Jacky Pai, Roham Pardakhtim, David Perek, Alexandre Poulin, Nick Raines, Salwa Rasheed, Leah Resneck, Kevin Rhodes, Ergys Ristani, Douglas Roberts\nw, Anik Sarker, Harrison Scarborough, Jonas Schmidtler, Brian Shieh, Benjamin Shuman, Garrett Simard, Forrest Smith, Udara Somarathna, Eliot Sosin, Jonathan Stewart, Kristine Strom, Adam Swett, Hal Talbot-Mantooth, Julia Tanner, Sai Sharan Thirunagari, Curtis Torrey, Simge Uzun, Eileen Wachs\nw, Rylan Wasson, Henry Wei, Jesse White, Beau Woodbyrne, Antonio Woods, Noah Young, Chandler Yount, Joseph Zhang, Sean Keller\\}%
    }%
  }%
  % Uncomment and fill in only if everyone shares this affiliation:
  \IEEEauthorblockA{\\Meta Reality Labs Research}
  \IEEEauthorblockA{\\}
  \IEEEauthorblockA{\ea Equal first authors to this work.}
  \IEEEauthorblockA{\nw Work done while at Meta.}
}
%IEEEauthorblockA{Affiliation TBD}}
\maketitle 

%\begin{strip}
%\centering
%\includegraphics[width=\textwidth]{figures/main.png}
%\caption{Fig. 1}{~High-level form-factor of our TACTUS digital gloves capable of high-fidelity tactile and pose sensing.}
%\label{fig:main}
%\end{strip}

\begin{abstract}
Collecting naturalistic human manipulation demonstrations with accurate contact force and hand pose is essential for robot dexterous manipulation policy and foundational world model learning---yet no existing wearable platform provides both calibrated fingertip force sensing and occlusion-resilient hand tracking in a form factor that preserves natural manipulation behavior. We present Tactus, a textile-integrated sensing glove that co-optimizes three design pillars: (1)~a thin, conformal 3D-knit form factor (\SI{1.5}{\milli\meter} fingertip stackup) designed to minimize interference with natural manipulation, (2)~strategically placed capacitive tactile sensors at manipulation-relevant regions informed by empirical contact studies, and (3)~IMU-fused pose tracking that maintains accuracy through the hand-object occlusions where manipulation data is most valuable. Tactus includes a documented calibration pipeline for cross-glove consistency and per-session calibration for subject-dependent correction. To enable objective comparison, we introduce standardized benchmarking protocols for tactile force regression and hand pose estimation, and apply them to Tactus and two commercially available tactile gloves — to our knowledge the first quantitative cross-platform evaluation of this kind. Tactus resolves a minimum detectable force of \SI{0.59}{\newton} at 95\% confidence, 1.6–2.7× lower than the commercial baselines, and with 0.87 cross-participant consistency, while our pose tracking sustains \SI{11.0}{\milli\meter} fingertip keypoint relative error during manipulation, when occlusion is heaviest. We show that these combined signals support object property estimation, slip detection, and representative recording of hand-object interaction — yielding demonstration data with the fidelity required for contact-aware manipulation policies and world models.

%Tactus includes a documented calibration pipeline for cross-glove consistency and per-session calibration for drift compensation---achieving 2.6\% relative force error with per-session calibration versus 51\% with end-of-line calibration alone. 

%To evaluate its performance, we introduce standardized benchmarking protocols about tacile force regression and hand pose estimation, and apply them to Tactus and commercially available gloves, establishing the first quantitative cross-platform comparison for tactile sensing gloves. Tactus achieves \SI{60.4}{gF} minimum detectable force at 95\% confidence (vs.\ \SI{95.7}{gF} for JQ, \SI{161.2}{gF} for PPS), 0.87 cross-participant consistency, and sub-\SI{100}{\milli\second} temporal response, while our best hand pose tracking achieves \SI{6.4}{\milli\meter} keypoint error while performing manipulation tasks. We demonstrate multi-sensory capabilities of the glove with object property estimation, slip detection, and hand-object interaction data recording.
\end{abstract}


%===================================================================
\section{Introduction}
%===================================================================
Learning dexterous manipulation from human demonstrations is a promising path toward general-purpose robot skills, and the quality of that demonstration data has been shown to directly impact downstream policy performance~\cite{lin2024datascaling}. There are three major options for capturing human manipulation sequences: egocentric-Computer-Vision (ego-CV) only; CV with an exoskeleton-based solution; and CV with a sensing-glove-based solution.

The ego-CV-only approach is the least intrusive of the three and obtains the most natural manipulation data, but CV cannot recover high-fidelity hand pose under occlusion, which becomes more severe exactly when hand-object interaction happens. Paliwal \textit{et al.} reported that in ego-CV recorded manipulation tasks, $<$ 10\% of the raw data eventually are usable for training purposes \cite{paliwal2026doasido}. Besides limitation in measuring hand pose, vision-based methods cannot directly measure the contact forces that are important for fine dexterity tasks. Exoskeleton-based systems like MILE~\cite{mile2025} provide high-fidelity kinesthetic data but constrain finger motion and alter grasp strategies; recent systems such as WHED~\cite{zhu2026whed} and DexEXO~\cite{zhu2026dexexo} achieved some progress, but still made great trade-off of naturalism with data quality. Sensing gloves promise both naturalism and direct force measurement, yet as we detail in the following section, no existing platform jointly provides calibrated force sensing, occlusion-resilient pose tracking, and a form factor validated to preserve natural grasp behavior.

\begin{figure}
    \centering
    \includegraphics[width=0.48\textwidth]{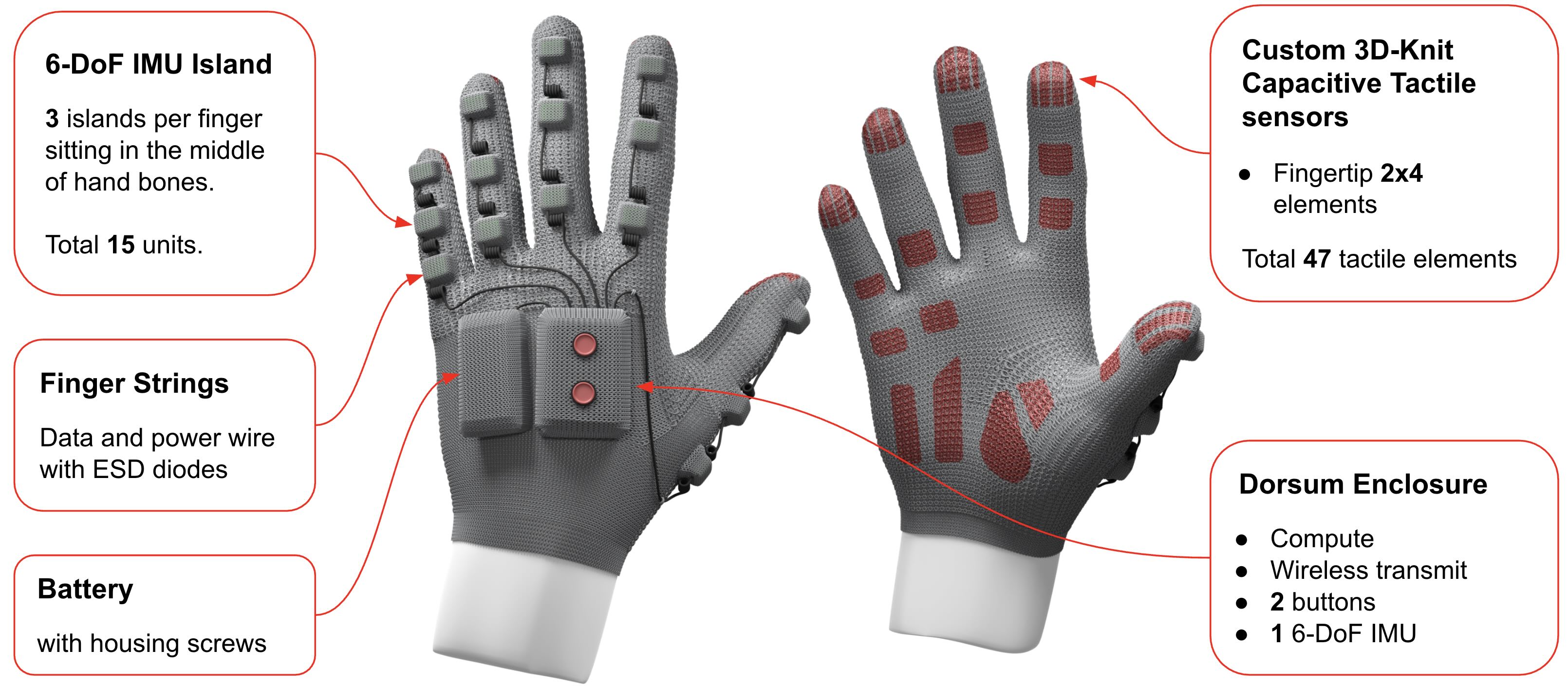}
    \caption{Form-factor of Tactus digital sensing gloves capable of high-fidelity tactile and pose sensing for natural hand-object manipulation data collection.}
    \label{fig:main}
\end{figure}

We present Tactus, a textile-integrated tactile and pose sensing glove designed from the ground up for naturalistic manipulation data collection, as shown in Fig. \ref{fig:main}. It is a 3D-knit textile glove, built in left- and right-hand variants, carrying 47 capacitive textile force sensing locations spanning fingertip to palm and 16 6-DoF IMUs; Sec. \ref{sec:electronics} details the sensing and electronics. Besides the glove design, our major contributions are the following:

\begin{itemize}
    \item \textbf{Minimal interference with natural grasp behavior, validated with human manipulation studies (Sec \ref{sec:task_coverage}).} A thin, conformal 3D-knit base glove (\SI{1.5}{\milli\meter} fingertip stackup) with low-profile, stretchable embedded sensors minimizes encumbrance while preserving the fidelity of natural manipulation.

    \item \textbf{Data-driven tactile sensor placement (Sec.~\ref{sec:tactile_design}) and efficient calibrated force sensing at every contact location (Sec.~\ref{sec:tactile_calibration}).} Placement is optimized by ink studies and a sensor-compliance threshold that maintains 65\% fingertip coverage across a broad anthropometric range in a single glove size. Calibrated forces are recovered in two stages: a scalable pneumatic-pressure factory calibration and a per-session online procedure.

    \item \textbf{Standardized benchmarks.} We introduce \textbf{Protocol-Tactile} (normal contact force \textbf{Sec. \ref{sec:force_benchmarking}}) and \textbf{Protocol-Pose} (hand pose tracking \textbf{Sec. \ref{sec:pose_benchmarking}}) and apply to both Tactus and commercially available gloves, yielding the first reproducible quantitative cross-platform comparison.

    \item \textbf{Application validation (Sec. \ref{section:application}).} Tactus tactile and pose data enables force-dependent robotic perception and manipulation tasks---object material classification, slip detection, and multi-sensory hand object manipulation.
\end{itemize}

%===================================================================
\section{Related Work}
%===================================================================

\textbf{Vision-Only Capture with Indirectly Sensed Contact / Force.} CV-only human hand object manipulation data capturing infers contact, and increasingly force, from imagery alone, trading traceable measurement for unencumbered, in-the-wild scale. Hand pressure has been regressed from a single RGB image using appearance cues such as skin blanching and soft-tissue spread~\cite{grady2022pressurevision}, extended to uncontrolled imagery~\cite{grady2024pressurevisionpp} and to the egocentric viewpoint~\cite{zhao2025egopressure}, while contact-only capture matured through physical proxies such as thermal residue~\cite{brahmbhatt2020contactpose}. Recent work moves from planar pressure to volumetric hand--object interaction, jointly estimating dense per-vertex contact and 3D force on hand and object meshes with force supervision synthesized in simulation~\cite{ilic2026egophi}, annotating in-the-wild egocentric clips with dense 3D contact correspondences~\cite{bansal2026epiccontact}, and targeting full-hand grasp pressure for everyday objects~\cite{zeng2026egotactile}. While these approaches are least intrusive and capture the most natural manipulation, their limitations are also obvious: their signals are \emph{indirectly sensed}---predictions supervised by simulation, by instrumented surfaces that constrain what can be manipulated, or by proxies that recover contact but not magnitude---so a calibrated, occlusion-invariant, worn measurement stream is precisely the supervision and evaluation signal they lack.

\textbf{Sensing Wearables for Manipulation Data Collection.} Wearable tactile sensing for manipulation data has evolved along two axes: spatial coverage and force measurement fidelity. The STAG glove~\cite{sundaram2019stag} achieved unprecedented coverage with 548 pressure sensors across the full hand, but without per-sensor calibration its force estimates remain unreliable; the OSMO platform~\cite{osmo2025} captures finger--object interaction with 12 magnetic sensors, again with no calibration pipeline; and commercial gloves from PPS~\cite{pps_glove} and JQ~\cite{jq_glove} offer fingertip force sensing in production-ready form factors but their performance on real manipulation tasks is not well-documented. The downstream value of calibrated tactile data is nonetheless established: Feel the Force~\cite{feeltheforce2025} achieved zero-shot transfer of human tactile demonstrations to robot policies, and MILE~\cite{mile2025} reduced both tracking and tactile error with an exoskeleton --- at the cost of constraining the very finger motion and contact whose grasp strategies it seeks to capture.

\textbf{Hand Pose Tracking for Manipulation.} Hand tracking during manipulation faces an inherent tension: the moments of greatest scientific interest---hand--object interaction, bimanual coordination, power grasps---are precisely when vision-based tracking degrades due to occlusion. Monocular and multi-view estimators~\cite{pavlakos2024reconstructing, han2022umetrack} have advanced rapidly, yet their evaluation is dominated by occlusion: hand--object benchmarks~\cite{arctic, hot3d} were built to expose exactly this regime, and the multi-method challenge report of Fan et al.~\cite{fan2024benchmarks} shows that no submitted method resolves hand--object interaction reliably under egocentric viewing. Marker-based mocap supplies the ground truth for these datasets and for corpora such as GRAB~\cite{grab}, but requires an instrumented volume and collapses when many markers are simultaneously invisible across all views, for which added modalities are the likely remedy~\cite{dextercap}. Gloves instead measure articulation directly and are occlusion-invariant by construction---via resistive or capacitive strain~\cite{glauser2019}, electromagnetic fields~\cite{saggio2025}, or helical stretch yarns with IMUs~\cite{tashakori2024}---but each modality carries a characteristic error source (strain hysteresis and per-wearer recalibration, field distortion, inertial drift and magnetic interference)~\cite{kim2020imu}, and validation studies find accuracy and repeatability differing markedly across technology and joint~\cite{henderson2021reliability}, which is largely why glove-derived pose is rarely used as a quantitative reference. Sparse-IMU fusion, well established for full-body pose~\cite{transpose}, has been extended to hands with inertial-only estimators on instrumented gloves~\cite{anik26} and then with learned cross-sensor attention that re-weights egocentric visual features against on-glove IMU signals as occlusion varies~\cite{kou2026aviht}. Tactus adopts this fusion formulation and places it on the same garment as calibrated tactile sensing, so that pose and contact force are captured in a single, co-registered stream.

\textbf{Tactile Sensing and Calibrations.} Wearable system force calibration remains under-studied. Exoskeleton trades manipulation naturalism with signal consistency. Calibrating a soft, stretchable tactile glove is challenging because of its well-documented strain artifact~\cite{jiang_capturing_2024}, especially in the worn state, where fit, grounding, and hand pose perturb the transfer function~\cite{park_finger_2019}. So existing work treats glove and wearer as an inextricable system and absorbs session variability through online calibration~\cite{osmo2025, zhang_dual-stage_2025, luo_learning_2021}, or end-to-end learning without any force reference~\cite{sundaram2019stag, song_opentouch_2025}. To date no method characterizes full-glove tactile performance without confounding from the wearer. A different approach loads all elements simultaneously with a known, spatially uniform pressure. A pneumatically controlled bladder with a thin elastic membrane conforms to the \textit{Device Under Test} (DUT), and sweeping the applied pressure yields per-sensor transfer functions. It shows promise on robot skins~\cite{kangro_plenum-based_2018}, insoles and other wearables~\cite{sorrentino_novel_2020}, and commercial sensor arrays~\cite{pps-calibration-white-paper}, but only on rigid or planar objects; uniform pressurization of a soft glove with inherently 3D, non-developable curvature is unaddressed, and is what our factory calibration method provides.

%===================================================================
\section{System Design}
%===================================================================
Our design choice is driven by a single requirement: capturing naturalistic manipulation demonstrations with quantitatively reliable force and pose data. While all gloves alter the pose and contact forces used in object interaction \cite{bensel1993}, we isolated three properties that can be tuned to \textit{minimize} such alteration to preserve naturalistic hand-object interactions: thickness, stiffness, and friction.

%Our design choice is driven by a single requirement: capturing naturalistic manipulation demonstrations with quantitatively reliable force and pose data. This requirement is in tension with instrumentation itself: every glove alters the poses and contact forces used in object interaction~\cite{bensel1993}. We term this degradation \textit{encumbrance}, and quantify it later in (Sec.~\ref{sec:task_coverage}). Encumbrance cannot be removed, only minimized, and we isolate three glove properties that govern it in this section.

%The boundary between tolerable and intolerable grasp delta is whether the user can maintain their natural pinch grasp type (2–3 fingers) with only a grip-force increase.
%\subsection{Form Factor for Natural Manipulation}
%The glove should maximally preserve the wearer's manipulation behavior, including the pose the hand takes and the forces it applies---this will provide the highest fidelity human action policies for robots to learn and adapt to.

\textbf{Thickness.} Task completion time increases approximately linearly with glove thickness~\cite{bensel1993}. Our task coverage evaluation (Protocol Dexterous-Tasks, details in Sec. \ref{sec:task_coverage}) also confirmed this relationship: glove configurations exceeding \SI{1.5}{\milli\meter} fingertip thickness produced measurable increases in task time and technique adaptations. Tactus (\SI{1.5} {\milli\meter}) reached similar barehand task completion time, best among all tested sensing glove candidates, thanks to our sensor design, which was knitted directly into the glove fabric---eliminating the added layer of discrete sensor laminated over or sandwiched between an existing textile stackup. 

\begin{figure}[t] \centering \includegraphics[width=0.45\textwidth]{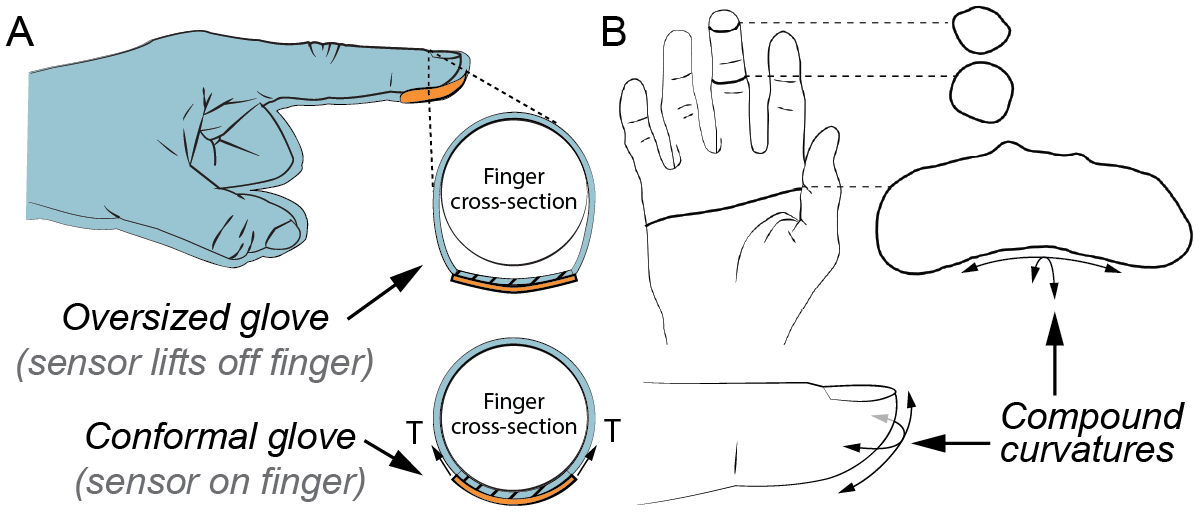} \caption{Glove fit sets whether sensor contact means hand contact. A) An oversized glove lifts the sensor off the skin, so it touches first. A conformal glove tensions sensors against the hand. B) Compound curvature demands sensor multi-axial conformation.} \label{fig:conformance} \end{figure}

\textbf{Stiffness.} As shown in Fig. \ref{fig:conformance}-B, low bending and tensile stiffness allow a glove to stretch around compound finger curvatures. Conformal fit preserves natural grasp behavior and ensures sensors remain coupled to the hand surface during contact---oversized gloves contact the environment before the hand does, and gloves that are either too big or too small increase task time by up to \SI{10}{\%}~\cite{size_ref}. The Tactus sensor achieves \SI{0.20}{\newton/\centi\meter} bending stiffness and \SI{0.22}{\newton/\centi\meter} tensile stiffness per \SI{25}{\milli\meter} sample width, conforming to the hand and tracking it through flexion. In contrast, PPS glove \cite{pps_glove} necessitates an oversized fit in which the sensor stands off from the hand surface (Figure \ref{fig:conformance}-A). This stiffness-fit coupling is reflected in encumbrance outcomes: Tactus achieved correspondingly lower NASA TLX load (smaller the better) \cite{hart2006nasatlx20} (2.97 vs 3.10) (full details in section \ref{sec:task_coverage}).

\textbf{Friction.} Surface friction must be high enough to maintain stable grasp without excessive grip force, yet low enough to permit the micro-slips essential for in-hand manipulation, such as pivoting and re-orientation \cite{johansson1984, adams2013}. The human skin coefficient of friction (COF) varies widely (0.4-1.0) against different surfaces \cite{adams2013}, and grip force is continuously modulated to maintain a safety margin above the slip threshold \cite{johansson1984}. Tactus design achieves sufficient interface friction through thermoplastic polyurethane (TPU) surface treatment in the fingertip regions, ensuring 0.4 COF (static friction on polished steel), a moderate value where stable grasp and slip prevention are achievable for dexterous manipulation.

%For the rest of this section, we focus on tactile sensing, pose sensing, and supporting electronics designs.

\subsection{Tactile Sensing Design}
\label{sec:tactile_design}

%The Tactus tactile sensor is a capacitive pressure sensor knitted directly into the textile substrate as a double-layer weft knit. The 3D-knit engineered textile is made of \SI{200}{dtex} spandex-core yarn, double-covered with polyamide~6.6, giving a low-profile, fully embedded sensor. Each capacitive cell consists of a world electrode (half-gauge knit conductive yarn) and a shield layer (interlock construction with two electrodes separated by high-stretch yarn); its capacitance varies with applied normal force. 

\begin{figure}[t] \centering \includegraphics[width=0.45\textwidth]{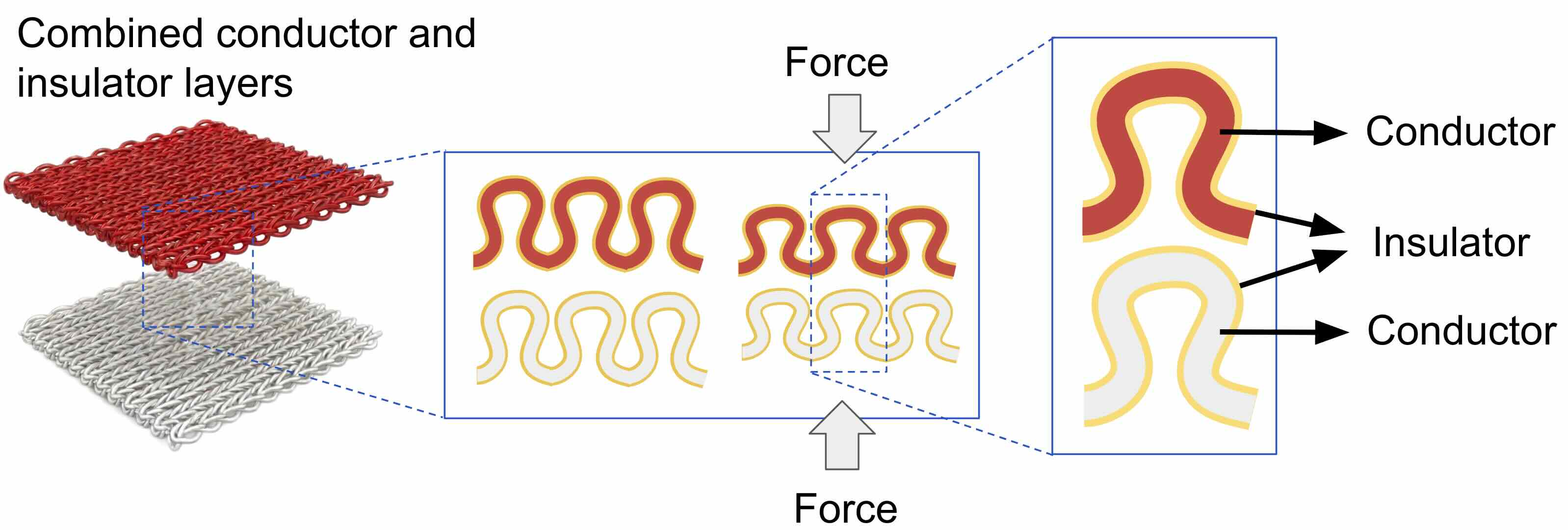} \caption{Knitted tactile capacitive sensor mechanism. The conductor and insulation layers form a capacitor whose capacitivity changes when compressed by external force.} \label{fig:ET} \end{figure}

The Tactus tactile sensor is a capacitive pressure sensor knitted directly into the textile substrate as a double-layer weft knit, giving a low-profile, fully embedded sensor, as shown in Fig. \ref{fig:ET}. The electrodes are custom insulated conductive wires, machine-knittable on industrial equipment, with knit structure chosen to meet sensor performance requirements: each capacitive cell pairs a half-gauge-knit upper electrode with an interlock lower layer, and its capacitance varies with applied normal force. 

We determine tactile sensor placement through three empirical studies: contact frequency mapping, hand geometry constraint analysis, and population coverage assessment. 

\begin{figure}[t] \centering \includegraphics[width=0.48\textwidth]{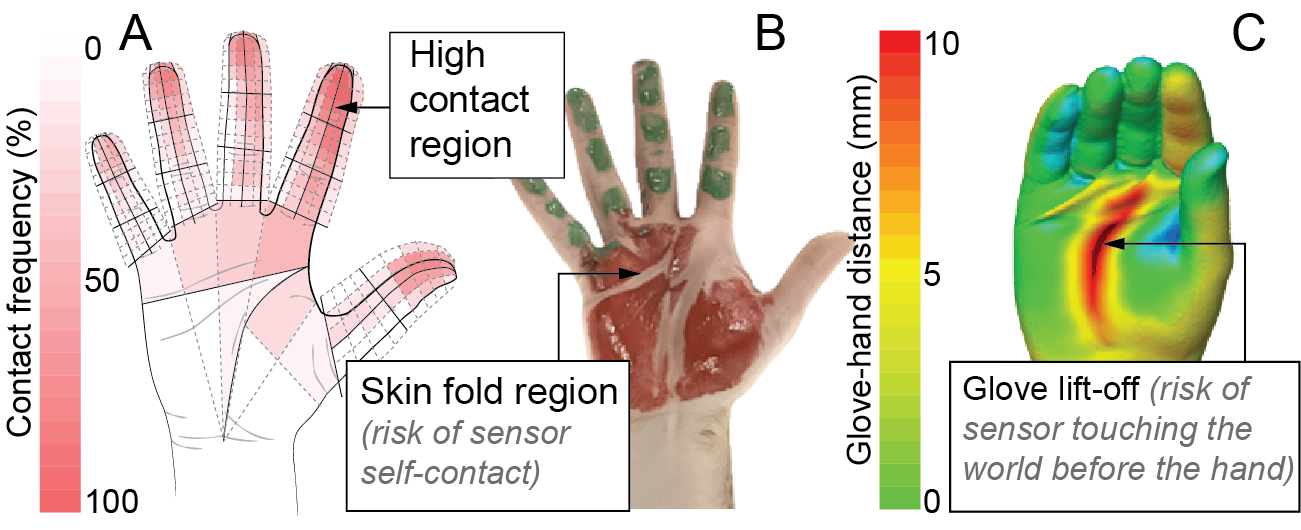} \caption{Sensor layout was informed by three datasets. A) A contact frequency map identifies where sensors are worth placing. B) Inking the flexed hand leaves skin folds un-inked, ruling out those regions because sensors there register self-contact. C) 3D scans of gloved- \& bare-hands rule out glove tenting regions, e.g. the palm center, where sensors contact the world first.} \label{fig:ink} \end{figure}

\textbf{Contact frequency.} An ink transfer study ($N{=}3$, 13 manipulation tasks) mapped contact frequency across hand regions (Fig.~\ref{fig:ink} A). Contact favored the radial side for index-pinky fingers and ulnar side for the thumb rather than the centerline. These frequencies establish coverage targets that minimize device complexity without sacrificing tactile data collection. 

\textbf{Hand geometry constraints.} Sensors cannot occupy all high-contact regions. Inking fully flexed hands ($N{=}6$) mapped fold lines across the finger and palmar surfaces; these compress during flexion and generate forces indistinguishable from object contact, so sensors must not span them (Fig.~\ref{fig:ink} B). Additionally, 3D scanning of gloved and un-gloved hands across multiple postures located palmar concavities (thenar and hypothenar valleys), where the glove bridges the hand surface rather than maintaining sensor--skin coupling, producing dead zones that cannot register force even under load (Fig.~\ref{fig:ink} C). Both define exclusion zones that refine the contact-driven coverage targets into specific sensor footprints. 

\begin{figure}[t] \centering \includegraphics[width=0.45\textwidth]{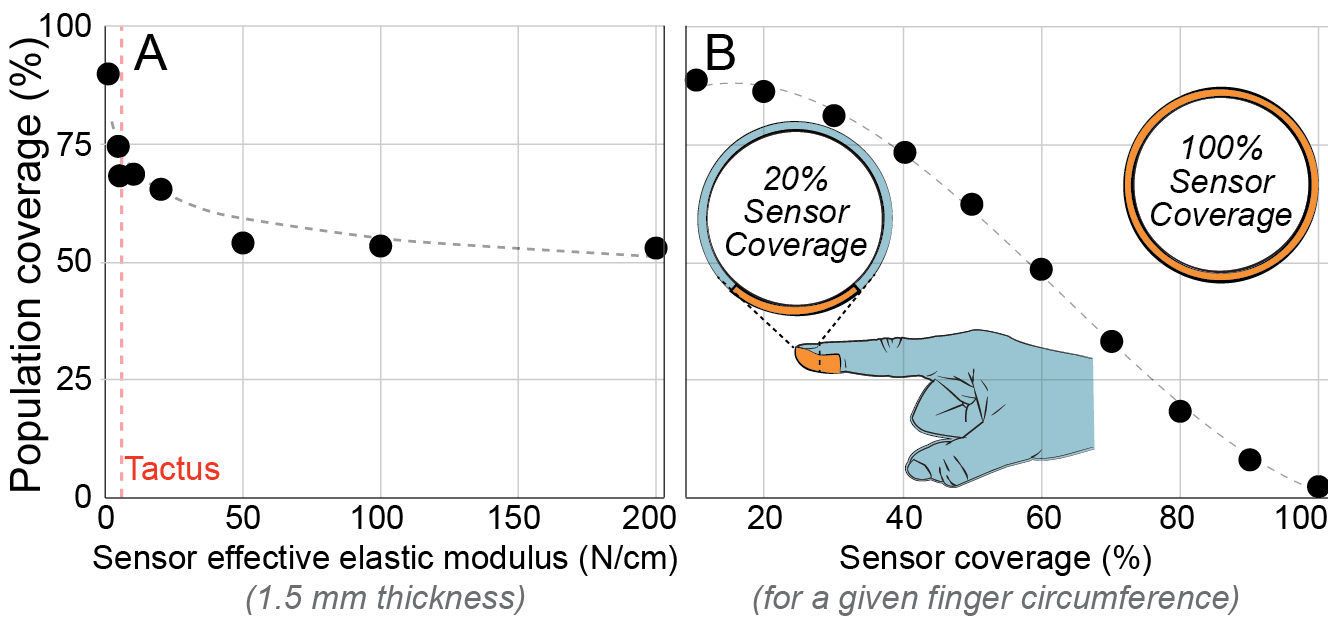} \caption{Two ways to widen population coverage. A) Increase the elasticity of the sensor. B) Reduce the sensor coverage area, which helps when the sensor is less stretchable than the base glove.} \label{fig:popcoverage} \end{figure}

\textbf{Population coverage.} A single glove size must accommodate a broad anthropometric range without sacrificing contact coverage. Population analysis established the threshold: sensors with an effective elastic modulus above \SI{35}{\newton / \centi\meter} (thickness excluded from calculations) behave as effectively inextensible in a glove system (Fig.~\ref{fig:popcoverage}). The Tactus sensor (approximately \SI{7}{\newton / \centi\meter}) remains well below this threshold, preserving glove elasticity through the sensing region (Fig.~\ref{fig:popcoverage} A). Sensor footprints are further constrained to less than 50\% of the finger circumference (Fig.~\ref{fig:popcoverage} B), leaving the remainder free to stretch and recover with a surrounding base textile made of high-denier elastane yarns for maximum compliance and elastic recovery, preventing the glove from bagging over repeated use. Together these choices maintain 65\% fingertip coverage across a broad population with one discrete glove size, whereas effectively inextensible sensors must limit coverage to approximately 40\% or require additional sizes. 

Based on the above studies, we finalize our tactile sensor placement as in Fig. \ref{fig:main}, each distal phalange carries a $4{\times}2$ array (8 independent sensing elements), while other phalanges carry at most single-element ($1{\times}1$) sensors. The total system comprises 47 sensor locations. In addition, a \SI{12}{\pico\farad} reference capacitor improves ADC resolution, \textit{i.e.} the number of discrete output codes an analog-to-digital converter can produce and reduces \textit{minimum detectable force} (MDF).

\subsection{Pose Tracking Design}
\label{sec:pose_design}

%IMU-based hand tracking has proven to be a viable path when combined with ego-CV view \cite{anik26}, especially in heavy occlusion scenarios, such as hand-object interaction, which are challenging for ego-CV tracking solution alone. We inherit the base design as in \cite{anik26}, and replaced the mid-phalange haptic island with an additional IMU island.

We aim to address the visual occlusion limitation of the ego-CV-only hand tracking solutions \cite{han2022umetrack, pavlakos2024reconstructing} during hand-object manipulation demonstrations for robotic tasks. Similar to \cite{anik26}, we use distributed IMUs that are invariant to visual occlusion, fused with egocentric vision so that each modality covers the other's failure mode: vision supplies absolute position and a drift-free global reference, while IMUs keep updating joint angle estimates during visual occlusions. 

Three principles shape the layout. First, every phalange is instrumented so that each finger joint angle is directly observable. Second, all units are 6-axis (accelerometer and gyroscope) rather than 9-axis: omitting the magnetometer removes the failure mode when approaching ferrous objects or active magnetic fields. Third, the IMUs are co-located with the tactile array board, rather than independently placed. The resulting configuration lands three 6-axis IMUs on each digit and one in the dorsum enclosure as a root reference (16 total), each sampling at \SI{200}{\hertz} (as in Fig. \ref{fig:main}). 

\subsection{Supporting Electronics} 
\label{sec:electronics}
Low encumbrance sensors require equally compliant supporting electronics---rigid wiring that cannot elongate with the finger restricts joint motion and negates the sensors' performance. Electrical routing addresses this encumbrance challenge by coiling conductors into a spring geometry: a mylar-wrapped, kevlar-core construction with TPU overmold, achieved ${\geq}70\%$ elongation at \SI{0.60}{\newton}, ${>}\SI{25}{\newton}$ failure load, and 168K cycles to 70\% elongation---providing the required stretch at a spring rate low enough to avoid perceptible resistance during natural finger motion.

All dorsum-mounted boards are housed in knit textile pockets rather than overmolded to the glove surface to preserve the surrounding textile's elasticity and prevent localized stiffening that would increase encumbrance. Each finger carries three phalange-level island PCBAs, each integrating a PSoC~6 microcontroller, an IMU (ICM-47688), and analog sensor connections, communicating over a UART finger bus. A dorsum-mounted dual-board controller stack provides WiFi+BLE communication, micro~SD storage (\SI{32}{GB+}), and USB-C (\SI{6.5}{Mbps}) for data download and charging. A \SI{900}{mAh} battery provides 3.5+ hours of untethered operation. Time-sync uses BLE or a SubGHz transceiver (S2-LP). 

%===================================================================
\section{Signal Modeling}
%===================================================================
\subsection{Tactile Calibration Pipeline}
\label{sec:tactile_calibration}

%Calibrated force measurement from a soft wearable sensor of a textile glove at the worn state is fundamentally harder than benchtop characterization. Donning introduces variable preload, the glove stretches differently across hand sizes, and forceful use changes sensor behavior over time.

\begin{figure}[t] \centering \includegraphics[width=0.45\textwidth]{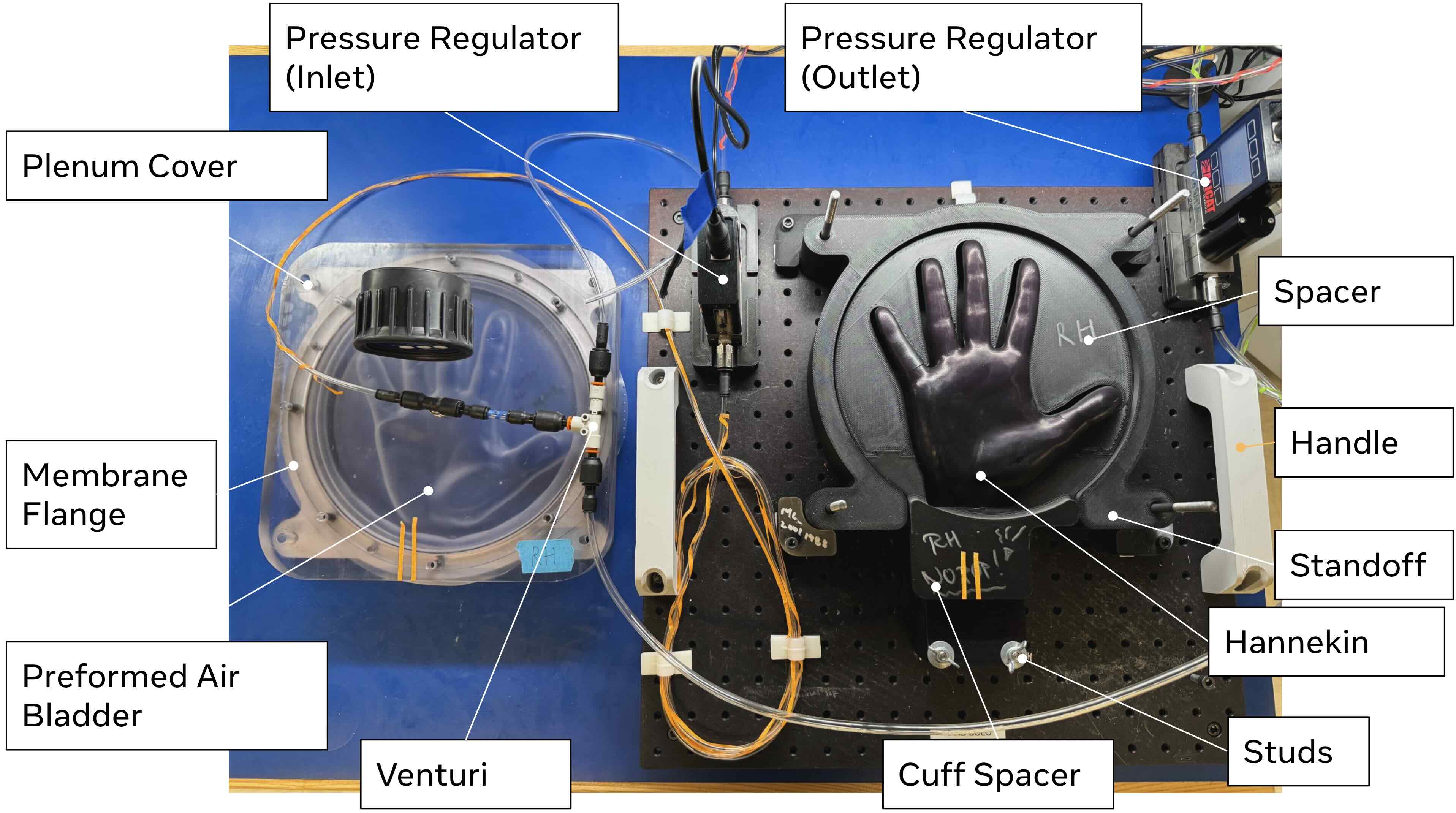} \caption{Factory calibration rig design. A closed pneumatic chamber in which the glove, donned on a prosthetic hand, is inflated against from above by a thermoformed elastomeric membrane with regulated air pressure.} \label{fig:factory_cal_rig} \end{figure}

%\textbf{Factory calibration rig.} We designed a calibration rig that sweeps all 53 glove sensor locations simultaneously, characterizing a complete glove in under \SI{23}{\minute}. As shown in Fig.\ref{fig:factory_cal_rig}, a membrane-based calibration system applies controlled, repeatable (${\ll}\SI{0.01}{\newton}$) pressure stimuli to sensors on the fully assembled glove donned on a prosthetic hand (``Hannekin''). Each sensor undergoes two pressure cycling sequences with $[0,15]$ PSIG in a step function to collect static response. An elastomeric membrane thermoformed with the hand shape ensures consistent palmar pressure distribution with ${\ll}\SI{0.1}{\newton}$ location-wise pressure variation. The system extracts per-sensor transfer functions using a cubic model $Y = A(X - B)^3$, along with other metrics including full span, sensitivity, hysteresis, noise, baseline, and fitting error ($R^2$) for sensor characterization.

\textbf{Characterization and factory calibration} Our calibration rig characterizes all 47 sensor locations of a complete glove simultaneously in under \SI{23}{\minute}. As shown in Fig.~\ref{fig:factory_cal_rig}, an elastomeric membrane thermo-formed to the hand shape presses against the assembled glove donned on a \textit{prosthetic hand}, applying controlled, repeatable (${\ll}\SI{0.01}{\newton}$) pressure stimuli with ${\ll}\SI{0.1}{\newton}$ location-wise variation; each sensor is driven through two $[0,15]$~PSIG step-cycling sequences. We then regress raw signal to pressure with a per-sensor cubic transfer function $Y = A(X-B)^3+C$, and extract full span, sensitivity, hysteresis, noise, baseline, and fitting error ($R^2$).

%\textbf{Drift characterization.} Longitudinal analysis of test gloves (used for over 15 hours of forceful data collection) versus an unused reference glove reveals two global drift mechanisms:
%\begin{itemize}
%    \item \textit{Don/doff sensitivity loss}: approximately 3\% reduction in full span per session, observed even on the unused reference. Attributed to viscoelastic fatigue from repetitive mechanical stretching.
%    \item \textit{Wearer-dependent hysteresis increase}: approximately 3\% per session on test gloves versus the reference. Attributed to inter-layer friction changes from physiological fouling (e.g. oil and sweat buildup).
%\end{itemize}
%Statistical analysis (one-sided Wilcoxon signed-rank test) found no robust evidence of localized sensor degradation from forceful use across most KPIs ($p > 0.05$). The exception is zero-load drift on two gloves ($p = 0.004$, $p = 0.04$), mitigatable by zero-offset taring.

With over 15 hours of forceful use of three Tactus gloves with periodic characterization, we find no evidence of localized sensor degradation: a one-sided Wilcoxon signed-rank test is null across most KPIs ($p>0.05$), the sole exception being zero-load drift on two gloves ($p=0.004$, $0.04$), which zero-offset taring mitigates. The drift we do observe is global and not attributable to sensor wear. Full span falls $\sim$3\% per don/doff session on test and reference gloves alike, consistent with viscoelastic fatigue from repeated stretching; hysteresis grows a further $\sim$3\% per session on test gloves only, consistent with inter-layer friction from physiological fouling. To address the glove-level sensor drift, along with don/doff variations and fit \& grounding differences, per-session calibration is needed with human-in-the-loop.

%\textbf{Per-session calibration necessity.} 

% We do observe that per-sensor transfer function can differ from session to session even for the unused reference glove. Compared to End-of-Line (EOL) factory calibration, per-data-collection-session calibration reduces pressure estimation error by 6--30 times. 

% Absent from additional human factors such as fit or pose variations, calibration provides a lower bound of force/pressure estimation error from dondoff variability alone, which can be attributed to location shift and grounding changes (e.g. bending radius and shape). This result indicates that one-time factory calibration is insufficient for tactile gloves, and per-session calibration is needed, especially with human-in-the-loop.

\textbf{Online per-session calibration.} Participants wear the glove and apply normal force from fingertip to a reference sensor. Force from each element is calculated by multiplying its area and pressure (using sensor output and factory calibration), and then aggregated over all fingertip elements to compare with the normal reference. To correct factory calibration, a scaling factor $\alpha$ is applied to the aggregated force, where $\alpha$ is derived from the best fit of predicted vs actual force $Y = \alpha X$. 

%Table~\ref{tab:human-cal} shows a reduction of 50.6\% absolute RMSE and 42.3\% relative error on average for force prediction with online per-session calibration compared to factory calibration, demonstrating that even a simple scaling factor can be effective for correcting errors introduced by human participants. 

As shown in Table~\ref{tab:human-cal}, a single per-session online scaling factor improves median force RMSE by $1.7$--$2.9\times$ and median relative error by $1.5$--$2.4\times$ over factory calibration alone. A one-parameter correction is thus sufficient to make reported forces comparable across gloves and sessions, which indicates that the dominant residual after factory calibration is a wearer-specific gain term --- fit, grounding, and hand pose --- rather than per-sensor error.

%\begin{table}[t]
%\centering
%\caption{EOL vs.\ Per-Session Calibration Accuracy (Calibration Rig)}
%\label{tab:cal}
%\begin{tabular}{@{}lcc@{}}
%\toprule
%Method & Median RMSE (PSI) & Median Relative Error \\
%\midrule
%EOL factory cal.      & 1.1--4.4  & 13\%--51\% \\
%Per-session cal.      & 0.17--0.30 & 1.6\%--2.6\% \\
%\bottomrule
%\end{tabular}
%\end{table}

\begin{table}[t]
\centering
\caption{Factory vs.\ Per-Session Online Calibration Accuracy}
\label{tab:human-cal}
\begin{tabular}{@{}lcc@{}}
\toprule
Method & Median full-glove force RMSE (N) & Median Relative Error \\
\midrule
Factory cal. &  2.50--5.22 & 40.6\%--98.1\% \\
Online cal. & 1.46--1.82 & 27.7\%--40.9\% \\
\bottomrule
\end{tabular}
\end{table}

%Absent from additional human factors such as fit or pose variations, calibration provides a lower bound of force/pressure estimation error from dondoff variability alone, which can be attributed to location shift and grounding changes (e.g. bending radius and shape). This result indicates that one-time factory calibration is insufficient for tactile gloves, and per-session calibration is needed, especially with human-in-the-loop.

%Factory calibration improved cross-glove $R^2$ consistency by approximately 17\% across 3~gloves in T1R benchmarking, enabling pooling data across glove units. 

\subsection{Hand Pose Tracking Pipeline}
\label{sec:pose_modeling}

The glove design enables different hand pose tracking modes. In this section, we highlight a few representative options, spanning the operating points a data-collection platform requires from real-time tracking for tele-operation to occlusion-agnostic offline solver feeding imitation policy training. Table \ref{tab:pose} in Sec.\ref{sec:pose_benchmarking} reports their tracking performance against marker-based mocap (OptiTrack) as the ground truth.

%\textbf{CV-only tracking.} A ResNet-based egocentric hand pose estimation network consumes Meta Quest~3 video streams and regresses a 26-DOF hand pose. The network uses the ResNet-based DPE framework~\cite{han2022umetrack} and is fine-tuned on both bare-hand and gloved-hand images, running \textbf{on-device} at real-time frame rate (30--60~Hz).

\textbf{CV+IMU Extended Kalman Filter (EKF) fusion.} We implement a ResNet-based ego-CV hand pose estimation network~\cite{han2022umetrack} that consumes ego-CV video streams and regresses a 26-DOF hand pose states, fine-tuned on both bare-hand and gloved-hand images. An EKF then fuses the on-glove IMUs, augmented with joint velocities, accelerations, and online-calibrated IMU-to-bone orientation parameters. Each update propagates state and covariance forward in time, then predicts expected sensor values from the current state and corrects the estimate against actual measurements (CV keypoints, IMU accelerometer and gyroscope). It consumes the CV and IMU observations asynchronously and runs in real-time.

%\textbf{Offline full-sequence solver.} A full-sequence non-linear optimization solver refines the hand trajectory by leveraging the complete temporal context. This method trades tracking accuracy with real-time capability, as a good candidate for post-data collection cleaning, and dataset curation.

\textbf{Adaptive vision-IMU Deep-Learning Fusion (AVI-HT).} A multi-modal deep learning network replacing the explicit EKF is implemented~\cite{kou2026aviht}. For each video frame, IMU samples are \textit{encoded} into tokens, which are fused with a global visual token through a hierarchical cross-sensor attention modules to implicitly model kinematic prior and local sensor relationship under occlusion. This is also a real-time solution.

\textbf{AVI-HT Offline Full-Sequence Solver.} A full-sequence non-linear optimization solver refines the hand trajectory provided by AVI-HT by leveraging the complete temporal context. This method trades real-time capability for tracking accuracy, as a good candidate for dataset quality curation.

%===================================================================
\section{Performance Evaluation}
%===================================================================

To assess whether Tactus supports dexterous manipulation data collection, we introduce three protocols measuring (A) how much the device interferes with natural hand manipulation task performance, (B) tactile-force sensing quality, and (C) hand pose estimation quality.

\subsection{Protocol Dexterous-Tasks} 
\label{sec:task_coverage}
To assess whether a sensing wearable preserves natural manipulation behavior, we conducted a within-participant encumbrance study ($N{=}5$) comparing barehand manipulation as baseline against our Tactus glove, a few state-of-the-art human manipulation tactile gloves (PPS glove \cite{pps_glove} and JQ glove \cite{jq_glove}), and one customized exo-skeleton device, as shown in Tab. \ref{tab:task_coverage}. For a total of 23 household, personal, and standardized dexterity tasks, we recorded task completion time ratio compared with barehand time, success rate, and adapted the NASA TLX task load \cite{hart2006nasatlx20}, which is a subjective, self-reported workload metric that combines six subscale ratings (mental, physical and temporal demand, performance, effort, and frustration) into one overall score on a 7-point Likert scale. 

%\begin{figure} \centering \includegraphics[width=0.45\textwidth]{figures/wearable_table.png} \caption{Task coverage evaluation candidates.} \label{fig:wearable_table} \end{figure}

%begin{table}[t] \centering 
%\caption{Wearable Task Coverage Performance} \label{tab:task_coverage} 
%\begin{tabular}{@{}lccc@{}} 
%\toprule 
%Wearable Type & \begin{tabular}[c]{@{}c@{}}Completion\\Time Ratio $\downarrow$ \end{tabular} & \begin{tabular}[c]{@{}c@{}}Task Success\\Rate $\uparrow$ \end{tabular} & \begin{tabular}[c]{@{}c@{}} NASA TLX\\Task Load \cite{hart2006nasatlx20} \end{tabular}\\ 
%\midrule 
%Barehand & 100\% & 100\% & 100\% \\ 
%PPS Glove \cite{pps_glove} & 100\% & 100\% & 100\% \\ 
%JQ Glove \cite{jq_glove} & 100\% & 100\% & 100\% \\ 
%Exoskeleton & 100\% & 100\% & 100\% \\ 
%Tactus Glove (ours) & 106\% & 100\% & 100\% \\ 
%\bottomrule 
%\end{tabular} \end{table}

\begin{table}
\centering
\caption{Wearables Task Coverage Performance vs bare-hand baseline}
\label{tab:task_coverage}
\setlength{\tabcolsep}{1pt}
\renewcommand{\arraystretch}{1.1}
\begin{tabular}{@{}lccccc@{}} 
\toprule
& \includegraphics[height=1.1cm,keepaspectratio]{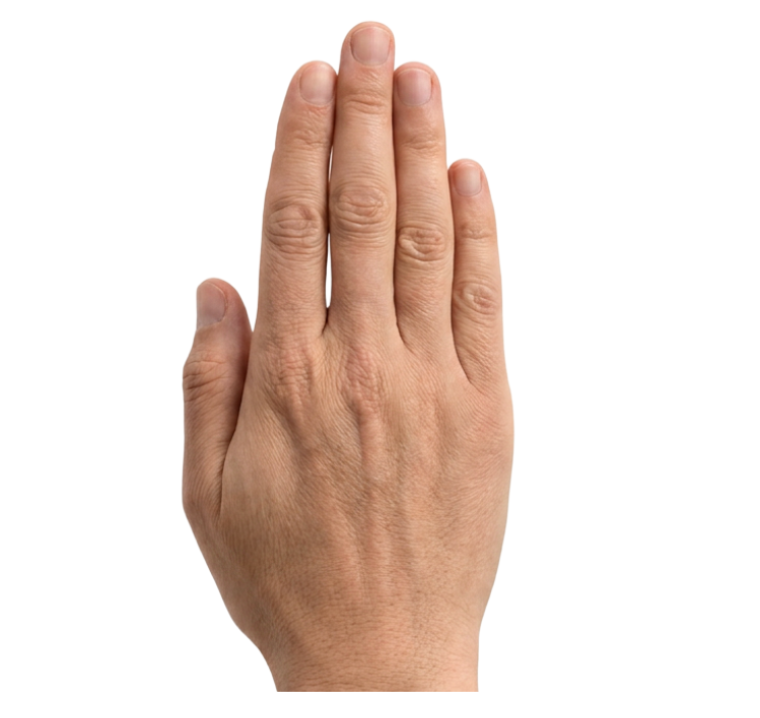}
& \includegraphics[height=1.1cm,keepaspectratio]{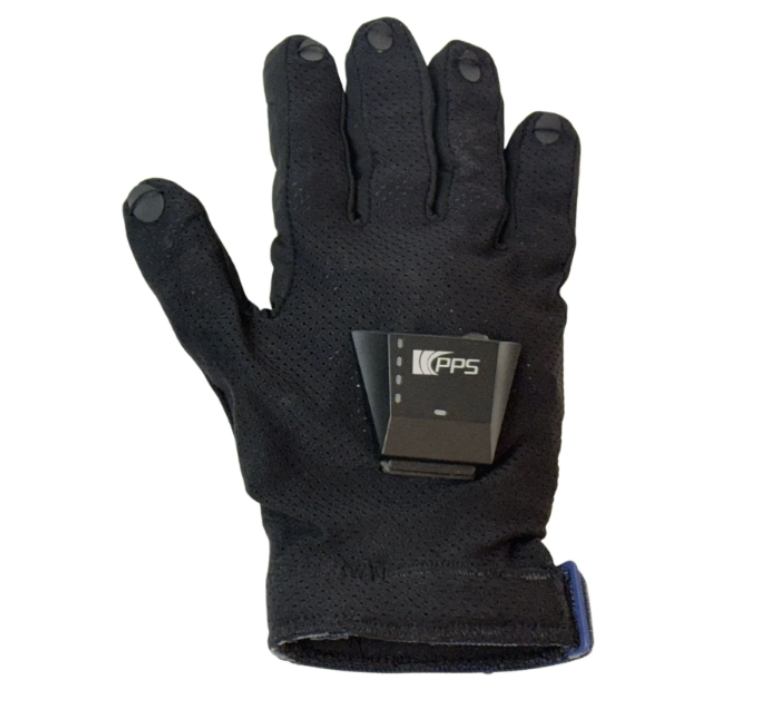}
& \includegraphics[height=1.1cm,keepaspectratio]{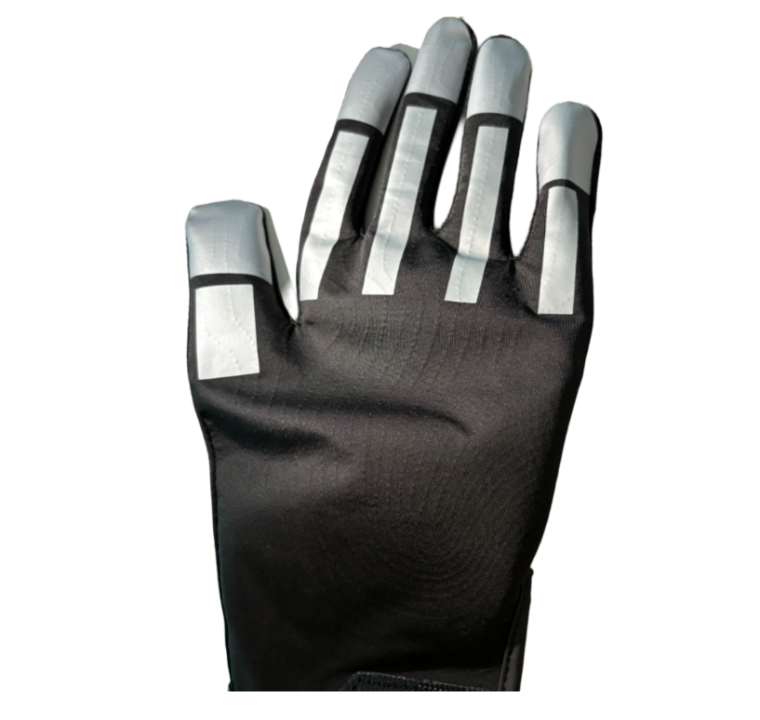}
& \includegraphics[height=1.1cm,keepaspectratio]{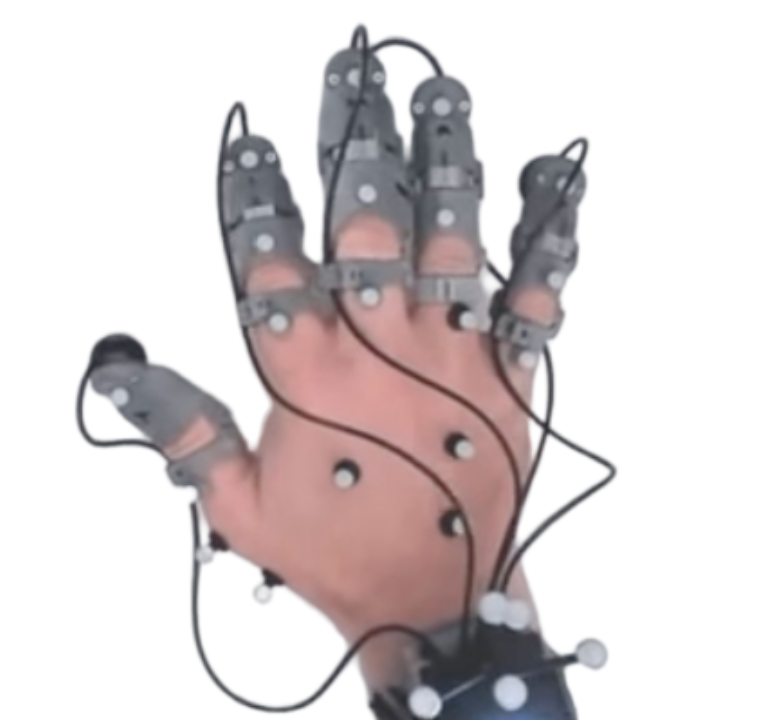}
& \includegraphics[height=1.1cm,keepaspectratio]{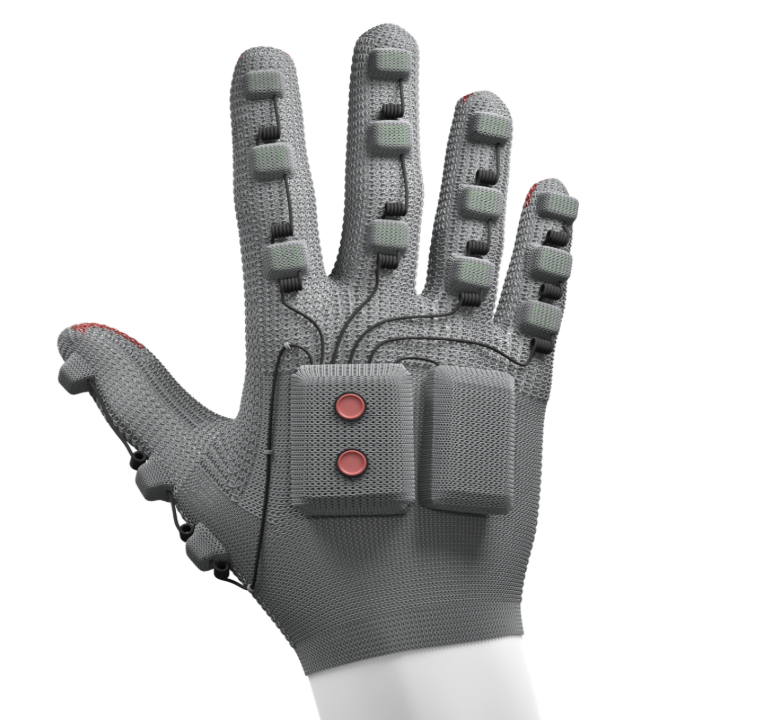} \\
& \begin{tabular}[c]{@{}c@{}}Barehand \\ (baseline)\end{tabular} & PPS & JQ & Exo & Tactus \\
\midrule
%\begin{tabular}[c]{@{}c@{}}\scriptsize{Completion} \\ \scriptsize{Time Ratio $\downarrow$} \end{tabular} & 100\% & 100\% & 100\% & 100\% & \textbf{106\%} \\
\scriptsize{Completion Time Ratio $\downarrow$} & 1.00 & 1.13 & 1.11 & 1.89 & \textbf{1.06} \\
\scriptsize{Task Success Rate $\uparrow$} & 100\% & 96.5\% & \textbf{100\%} & 95.7\% & \textbf{100\%} \\
%\scriptsize{NASA TLX \cite{hart2006nasatlx20} $\downarrow$} &  \begin{tabular}[c]{@{}c@{}}1.87\\($\sigma$~0.96)\end{tabular} & -- & TBA & TBA & \textbf{3.2} \\
\scriptsize{NASA TLX [1---7] $\downarrow$} & 1.87 & 3.10 & 3.43 & 5.23 & \textbf{2.97} \\
\bottomrule
\end{tabular}
\end{table}

%Across the three conditions the ordering was Barehand $>$ Tactus $\gtrsim$ PPS. Neither glove differed significantly from bare hand in task completion time, and Tactus and PPS were statistically indistinguishable from each other. Tactus matched bare hand on success rate (\SI{100}{\%}, vs.\ \SI{96.5}{\%} for PPS) and required fewer technique changes relative to bare hand (7 of 23 tasks, vs.\ 9 for PPS). Participants rated Tactus as most closely resembling natural hand interaction, which they attributed to its thin, low-profile knit construction with no protruding sensors; its principal weakness was surface slipperiness (addressed below). Two caveats apply: the sample is small ($n{=}5$), and the PPS glove used a thicker, denser base glove that is not a one-to-one construction match.

Tactus achieves the best performance among the tested wearables against all the three metrics, validating our assumption that our design achieves the closest resemblance to barehand object manipulation compared with other wearables. With almost the same task completion time and slight trade-off of subjective workload, what we gain is high quality perceptual information for hand-object manipulation data.

\subsection{Protocol Tactile}
\label{sec:force_benchmarking}

%\textbf{T1R (robot-based).} A robot end-effector applies controlled normal forces at predefined locations on the gloved hand form, isolating sensor performance from human variability.

%\textbf{T1H (human-in-the-loop).} 
Participants ($N{=}3$) perform varying-pose fingertip pressing tasks as shown in Tab.~\ref{tab:force} Left. We then train an MLP force regression model for each of glove candidates supervised by the 6DoF force-torque sensor (Nano-17), and report 5-fold validation about the sensor performance. The protocol captures ecological performance including don/doff variation and \textit{Pose-Related Artifacts} or PRA (changes in tactile sensor signals caused by strain from hand poses, rather than by contact force) \cite{yu2026PRA}. To measure that, we define \textit{PRA ratio} as the maximum per-channel ratio between the relative signal during in-air free motion and 3N normal load after the base signal load compensation, or ${{max}(|S_{in\_air\_motion} - S_{base}|)} / {{max}(|S_{3N\_load} - S_{base}|)}$. For an \textit{ideal} sensing wearable, its PRA ratio should be zero.

%\begin{equation}
%r_{PRA} = \frac{{max}(|S_{in\_air\_motion} - S_{baseline}|)}{{max}(|S_{300gF\_load} - S_{baseline}|)}
%\end{equation}

Results in Tab.~\ref{tab:force} Right show that comparing with PPS and JQ tactile sensing gloves, Tactus achieves the lowest Minimum Detectable Force (MDF), the highest cross-participant consistency, and the lowest PRA ratio, all thanks to its design choices and calibration process. The JQ glove achieves slightly better point accuracy (\SI{0.18}{N} vs.\ \SI{0.20}{N})---but neither JQ nor PPS provides a documented calibration pipeline to maintain performance across subjects or sessions. 

Note that the results here are not directly comparable with the calibration performance in Sec.~\ref{sec:tactile_calibration}, where Tab.~\ref{tab:human-cal} gives per-sensor cubic transfer functions across all 47 locations over the \textit{full pressure range}, while here Tab.~\ref{tab:force} gives a learned multi-channel MLP for \emph{fingertip} normal force at the lighter forces for glove-to-glove performance comparison purposes.

\begin{table}[t] \centering 
\caption{Tactile benchmark result. Left: testing rig, and pressing locations at the fingertip. Right: tactile gloves' force regression performance comparison.} 
\begin{minipage}{0.19\linewidth} \centering \includegraphics[width=\linewidth]{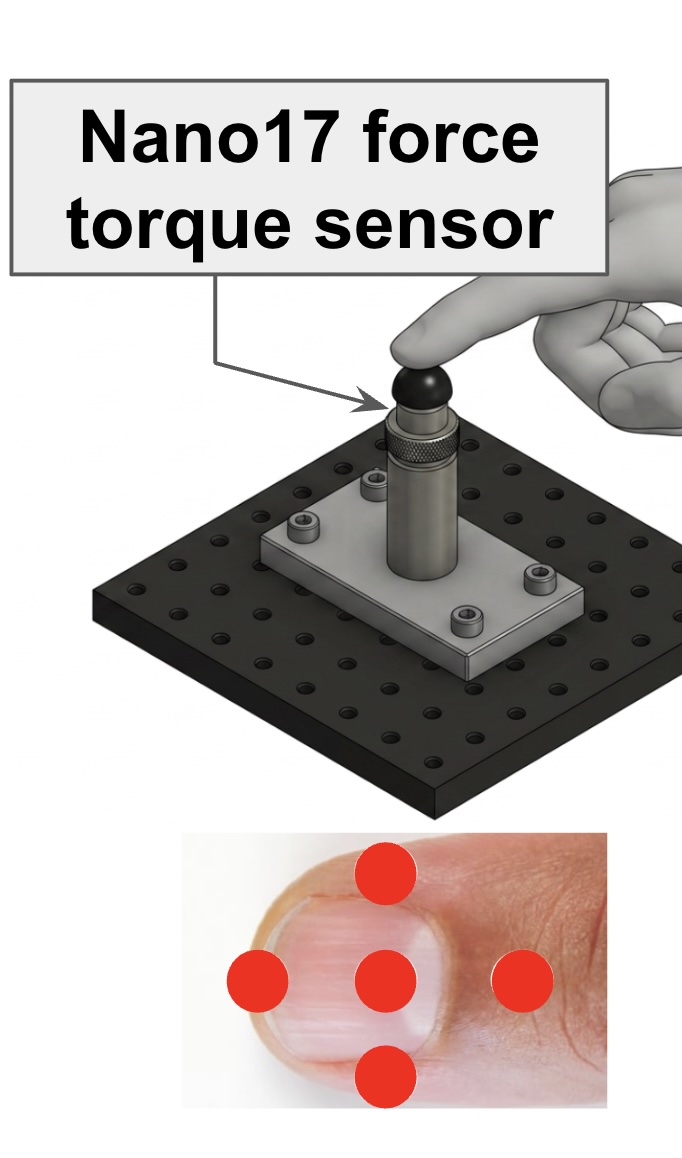} \end{minipage}\hfill \begin{minipage}{0.81\linewidth} \centering \begin{tabular}{@{}l ccc@{}} \toprule Metric & PPS & JQ & Tactus\\ \midrule \scriptsize{MDF@95\% (N)$\downarrow$} & 1.58 & 0.94 & \textbf{0.59} \\ \scriptsize{Fingertip force error (N)$\downarrow$} & 0.23 & \textbf{0.18} & 0.20 \\ \scriptsize{X-subj. consistency$\uparrow$} & 0.77 & 0.86 & \textbf{0.87} \\ \scriptsize{PRA ratio $\downarrow$}  & 2.13 & 0.43 & \textbf{0.17} \\ \bottomrule \end{tabular} \end{minipage} \label{tab:force} 
\end{table}

%\begin{table}[t]
%\centering
%\caption{Force Sensing Benchmark Comparison}
%\label{tab:force}
%\begin{tabular}{@{}lccc@{}}
%\toprule
%Metric & Tactus & JQ & PPS \\
%\midrule
%MDF @95\% T1H (gF)      & \textbf{60.4}  & 95.7  & 161.2 \\
%Accuracy T1H (gF)       & 19.9           & \textbf{18.0} & 23.0  \\
%Cross-subj.\ T1H        & \textbf{0.87}  & 0.86  & 0.77  \\
%PRA T1H                  & \textbf{121.8{\scriptsize$\pm$91.5}} & --- & 139.4{\scriptsize$\pm$151.1} \\
%\bottomrule
%\end{tabular}
%\vspace{-2mm}
%\end{table}

\subsection{Protocol Hand Pose}
\label{sec:pose_benchmarking}

%The T6 protocol evaluates hand pose tracking accuracy against OptiTrack ground truth (Table~\ref{tab:pose}). MKPE.T (root-aligned) isolates joint angle accuracy from absolute positioning error. The most significant gains from IMU fusion appear in relative metrics, confirming that IMUs primarily improve joint angle accuracy---the more relevant metric for manipulation analysis. The offline solver achieves the highest availability (97.5\%) among CV-dependent modes while maintaining the best absolute accuracy (\SI{13.4}{\milli\meter}), making it the recommended mode for post-hoc dataset curation. OptiTrack ground truth itself has 84.4\% marker availability, establishing an inherent ceiling for CV-dependent evaluation.

Three tracking modes as in Sec.\ref{sec:pose_modeling} are evaluated against an ego-CV bare-hand baseline UmeTrack \cite{han2022umetrack}. %We also add an off-the-shelf tracking Manus Quantum glove \cite{manus-quantum} widely-used in robotics teleoperation for comparison. 
The ground truth pose is obtained by attaching OptiTrack mocap markers to hand key points, and solving the inverse kinematic pose. Participants ($N$=5) each perform a fixed set of 16 object manipulation tasks spanning dexterous manipulation task space (e.g. YCB object manipulation \cite{calli2015ycb}). As shown in Tab.~\ref{tab:pose}, \textsc{mkpe} reports mean keypoint position error over the whole hand and \textsc{f.mkpe} restricts it to the fingertips; the \textsc{mkpe.t} and \textsc{f.mkpe.t} variants are computed after translation alignment at the wrist, isolating joint-angle accuracy from absolute positioning error. The transformed fingertip accuracy is the most relevant quantity for manipulation analysis, where relative finger geometry determines contact. 

The fused modes are all more accurate than the CV-barehand \cite{han2022umetrack} baseline. Adding on-glove IMUs to the causal vision estimate reduces root-aligned error by 11\% and root-aligned fingertip error by 17\%, while absolute error improves by under 3\%---confirming that the IMUs supply joint-angle information rather than global position, which vision already provides. AVI-HT inverts this profile: its largest gains are in the absolute metrics (34\% \textsc{mkpe}, 33\% \textsc{f.mkpe} relative to the vision-only baseline) while retaining the relative improvements, indicating that attention over per-sensor inertial features recovers global hand configuration that the explicitly-modelled filter does not. Full-sequence optimization improves all four metrics further by exploiting complete temporal context, making it the preferred mode for post-hoc dataset curation where latency is irrelevant.

%\todo{Add Manus result in the table and explain why the performance look like that.}

\begin{table}[!b]
\centering
\caption{Hand Pose Benchmarks Performance}
\label{tab:pose}
\begin{tabular}{@{}lcccc@{}}
\toprule
Tracking Mode & \begin{tabular}[c]{@{}c@{}}MKPE\\(mm) $\downarrow$ \end{tabular} & \begin{tabular}[c]{@{}c@{}}MKPE.T\\(mm) $\downarrow$ \end{tabular} & \begin{tabular}[c]{@{}c@{}}F.MKPE\\(mm) $\downarrow$ \end{tabular} & \begin{tabular}[c]{@{}c@{}}F.MKPE.T\\(mm) $\downarrow$ \end{tabular} \\
%& \begin{tabular}[c]{@{}c@{}}Jitter\\(mm)\end{tabular} \\
%& \begin{tabular}[c]{@{}c@{}}Lost\\(\%)\end{tabular} \\
\midrule
CV-barehand \cite{han2022umetrack}  & 15.8 & 8.9 & 19.8 & 17.1 \\ % & 0.94\\ % & 1.1 \\
CV+IMU EKF                  & 15.4 & 7.9 & 18.7 & 14.2 \\ %& 1.08\\ % & 1.7 \\
%CV+IMU offline          & 7.8  & 13.6 & \textbf{0.83} & 1.1 \\
AVI-HT \cite{kou2026aviht}  & 10.4 & 7.0 & 13.3 & 12.8 \\ %& 2.61\\% & 1.7 \\
AVI-HT offline              & \textbf{9.7}  & \textbf{6.4} & \textbf{12.1} & \textbf{11.0} \\ %& 2.26 \\% & 1.1 \\
%Manus Quantum              & ---  & 11.8 & --- & 21.7 \\
\bottomrule
\end{tabular}
\end{table}

%\subsection{Data Quality}
%The factory calibration longitudinal analysis demonstrates that Tactus sensors maintain performance over approximately 15 hours of forceful data collection: no robust evidence of localized degradation (Wilcoxon $p > 0.05$), membrane rig repeatability ${\ll}\SI{0.1}{\newton}$, and cubic model $R^2 > 0.9$ across sessions. These stability characteristics enable multi-day data collection campaigns with quantified confidence in measurement accuracy.
%\todo{Add a double-column figure about the performance of glove and its signal stream visualization.}

%===================================================================
\section{Applications: Force-Dependent Tasks}
%===================================================================
\label{section:application}

The benchmarks above establish that Tactus captures calibrated fingertip force and occlusion-resilient hand pose. We now present some capabilities that enable downstream manipulation tasks that indirect sensing cannot support---the core value of Tactus for high-fidelity, natural data collection.

\subsection{Surface Material Estimation}

Inferring object material properties provides critical context from hand-object manipulation, and is a direct test about the tactile signal fidelity. We evaluate a $9$-category \textit{roughness} classification on Tactus. As shown in Tab. \ref{tab:roughness}, given a set of 9 texture pads for children cognitive development, participants slide the index finger across each pad while the glove records the tactile signal, and we compare two models: a tactile-only network consisting of a stack of transformer modules followed by an MLP head, and a zero-shot vision--language baseline (Llama-3.2-11B-Vision-Instruct) that predicts the texture category from an egocentric image and a text prompt. The two models are expected to exhibit complementary strengths---the tactile network resolving fine surface structure that is not visually apparent, the VLM exploiting semantic priors about materials---which motivates tactile-augmented visual language modeling as a direction for object attribute inference.

\begin{table}[t]
\centering
\caption{Surface roughness 9-class accuracy table. The tactile network vs VLM zero-shot performance.}
\label{tab:roughness}
\setlength{\tabcolsep}{1pt}
\renewcommand{\arraystretch}{1.1}
\begin{tabular}{lcccccccccc}
\toprule
Model & Overall & Mirror & Bushy & Scaly & Bumpy & Braided & Ridged & Fuzzy & Fluffy & Hairy \\
\midrule
Tactile & \textbf{0.75} & 0.78 & \textbf{0.83} & 0.66 & 0.81 & \textbf{0.67} & 0.61 & 0.67 & \textbf{0.97} & \textbf{0.84} \\
VLM     & 0.56 & \textbf{0.84} & 0.00 & \textbf{0.96} & \textbf{0.97} & 0.08 & \textbf{1.00} & \textbf{0.86} & 0.45 & 0.00 \\
\bottomrule
\end{tabular}
\end{table}

\subsection{Slip Detection}
\begin{figure}[t] \centering \includegraphics[width=0.48\textwidth]{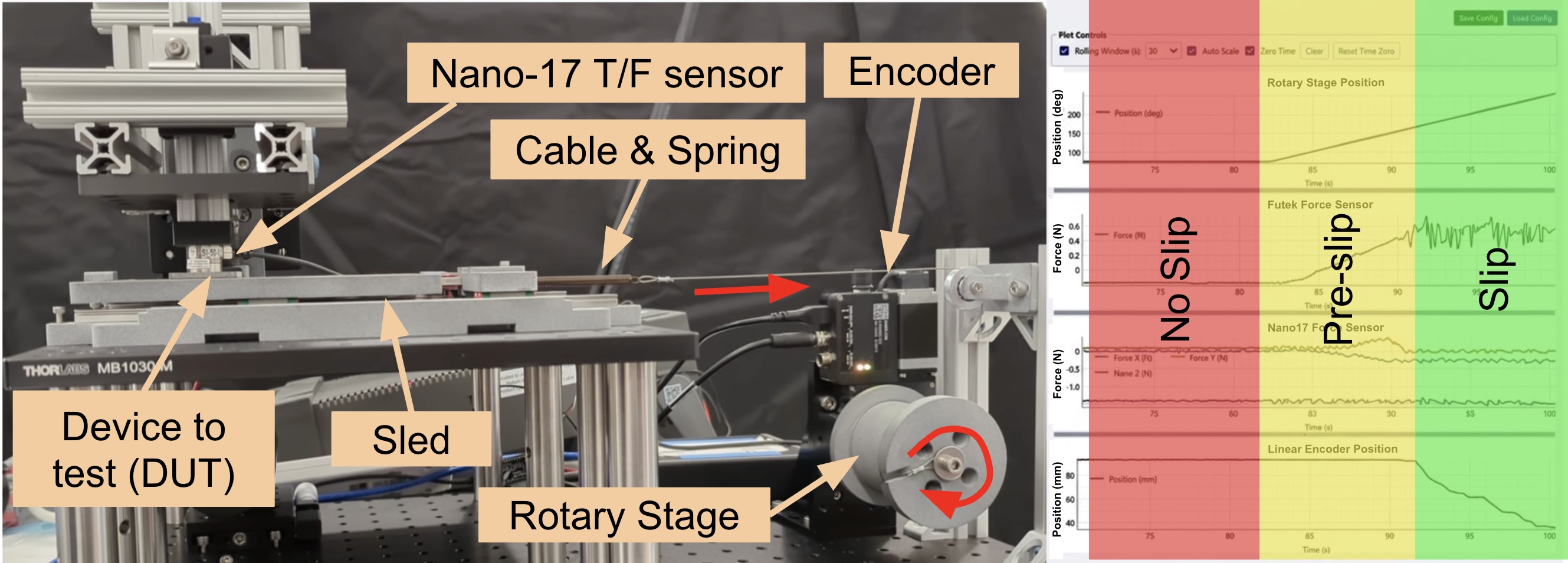} \caption{We design a rig to test controlled slip events with measured force and displacement. We can automatically label different stages of slip from the ground truth sensory signals.} \label{fig:slip_rig} \end{figure}
Detecting the onset of slip, i.e. when an object starts to move relatively to hand, is fundamental to stable manipulation. It requires the capability to detect rapid \textit{directional force transients}. As shown in Fig. \ref{fig:slip_rig}, a rotary stage pulls a sled via a cable and spring; the DUT sits on a tower above it. The spring ramps pull force smoothly until static friction breaks, then decompresses so the sled re-sticks, producing repeated natural stick-slip cycles rather than a single commanded shear. Ground truth labels (\textsc{no-slip}~/~\textsc{pre-slip}~/~\textsc{slip}) come from encoder position plus tangential load. 

%Across 7 (5 training, 2 testing) sessions ($\sim$34 slip events each), a compact causal temporal CNN consuming a 4\,sec window of the $4\times2$ fingertip pressure channels detected slip with $\mathrm{F1}=1.00$, firing within $0.01$\,sec of labelled onset. This means, although Tactus tactile sensors output pressure \textit{only} and perform no force decomposition, distinctive enough multi-channel signal profile resolves an event whose defining physics is the onset of tangential relative motion. We attribute this to the \textit{spatial distribution} of its array: as the contacting area shears, the pressure distribution deforms in a \textit{characteristic signal pattern}, so shear is represented implicitly even though not directly measured. One thing to point out though, the implicit-shear account remains under-determined, since a normal-pressure distribution does not uniquely specify shear: shear uniform along the load direction leaves no distributional signature, and a single-layer array cannot separate asymmetric normal loading from shear. There are follow-up future experiments to carry out.

Across 7 sessions (5 training, 2 testing; $\sim$34 slip events each), a compact causal temporal CNN consuming a 4\,sec window of the $4\times2$ fingertip pressure channels detected slip with $\mathrm{F1}=1.00$ on a single train/test split, firing within $0.01$\,sec of labelled onset. Although Tactus outputs pressure \textit{only} and performs no force decomposition, the multi-channel signal profile is distinctive enough to resolve an event whose defining physics is the onset of tangential relative motion. We attribute this to the \textit{spatial distribution} of the array: as the contact area shears, the pressure distribution deforms in a \textit{characteristic pattern}, so shear is represented implicitly without being measured directly. 

The account remains under-determined, however, since a normal-pressure distribution does not uniquely specify shear. We regard this only as a feasibility result; follow-up experiments across additional preloads, slip directions, and a second held-out group are planned.

\subsection{Object Manipulation Modeling}
\label{section:obj_manip}
We demonstrate on-glove pose and tactile signals fusion yields higher quality manipulation trajectories than egocentric vision solution alone---both the object and hand---and therefore more suitable for robotic policy learning. For our setup: a participant manipulates objects spanning a range of size, geometry, and grasp affordance, each with a pre-scanned mesh. Three streams record concurrently: the glove's IMU and tactile signals, an ego-CV RGBD camera. Streams are time-aligned. For each object, the participant performs a natural fingertip grasp, 3D translation and rotation, and free-form manipulation. The trials deliberately induce hand--object and hand--hand occlusion as would happen in natural manipulation.

\begin{figure}[t] \centering \includegraphics[width=0.485\textwidth]{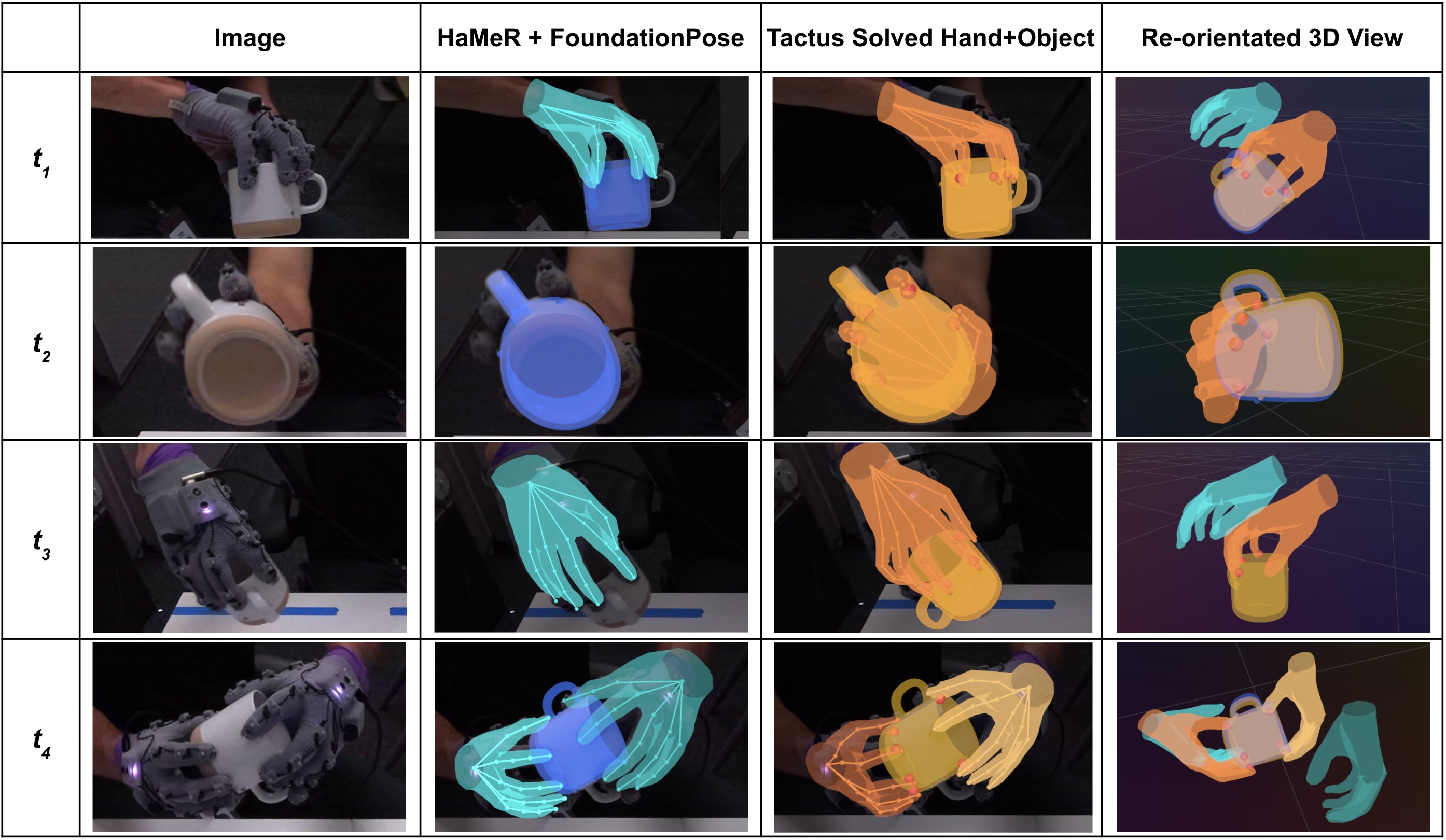} \caption{Multi-modalities from Tactus (Tactile and IMU) enable qualitatively better hand-object interaction recording compared to CV-only HaMeR hand pose and RGBD CV-only FoundationPose object pose. $t_1$: FoundationPose has hard time to pin-point the cup handle orientation for such close-to-symmetric object. $t_2$ \& $t_3$: HaMeR hand and FoundationPose object loses track completely due to occlusion. $t_4$: although HaMeR recovers both hands, the hands scaling / depth are incorrectly estimated. } \label{fig:hand-obj-tracking} \end{figure}

We compare against vision-only baselines---HaMeR \cite{pavlakos2024reconstructing} for hand pose and FoundationPose \cite{foundationposewen2024} for object pose qualitatively by overlaying the hand and object poses to the original video. Because contact constrains kinematics between the hand and object, we are able to achieve high-fidelity hand and object pose together, as in Fig. \ref{fig:hand-obj-tracking}. The tracking quality can be greatly improved when applying Tactus pose and tactile information to the joint optimization. 

%===================================================================
%\section{Limitations}
%===================================================================
%\input{sections/discussion}

%===================================================================
\section{Conclusion and Future Work}
%===================================================================
Tactus provides calibrated fingertip force sensing and occlusion-resilient hand tracking in a textile-integrated form factor that preserves natural grasp behavior. Its contribution is a complete measurement system: sensor placement derived from empirical contact studies, a documented calibration pipeline with quantified drift, and standardized benchmarking protocols that make cross-platform comparison reproducible.

Applied to Tactus and to commercially available gloves, these protocols establish quantitative performance bars: \SI{0.59}{N} minimum detectable force at 95\% confidence, \SI{11.0}{mm} root-aligned fingertip tracking error from offline trajectory optimization, at 106\% of bare-hand task completion time. These capabilities position Tactus as a platform for the force-dependent natural manipulation tasks driving the next generation of robot learning. As importantly, the introduced benchmarking protocols are the transferable artifact: by specifying ground truth, metrics, and drift reporting, they offer the field a route from incomparable single-platform demonstrations toward reproducible, quantitative evaluation of sensing gloves.

Our major future work is large-scale user study spanning more participants, surfaces, slip detection, and glove platforms under Protocol-Tactile and Protocol-Pose --- the protocols are specified precisely so that this extension is possible by others as well as by us. We would also like to reduce the per-session calibration burden as a parallel thread. We will also systematically report Tactus' performance degradation during the full life-cycle.

\bibliographystyle{IEEEtran}
%\begin{thebibliography}{10}
\bibliography{sections/reference}
%\end{thebibliography}
\end{document}